\documentclass[11pt,a4paper]{article}
\usepackage[utf8]{inputenc}
\usepackage{graphicx}
\usepackage[margin=2cm]{geometry}
\usepackage[acronym]{glossaries}
\usepackage[natbib=true,
sorting=none, doi=false,isbn=false,url=false,eprint=false,mincitenames=1,maxcitenames=1,uniquelist=false,alldates=year,maxbibnames=20]{biblatex} %Imports biblatex package
\usepackage{xcolor, soul}
\sethlcolor{pink}
\usepackage{url}
\usepackage[hidelinks]{hyperref}
\usepackage{ragged2e}

\newcommand{\cmt}[1]{}
\newcommand{\rvw}[2]{#2}

\usepackage{booktabs}
\usepackage{tabularx}
\usepackage[group-separator={,}]{siunitx}
\usepackage{caption}

\title{End-to-end Risk Prediction of Atrial Fibrillation from the 12-Lead ECG by Deep Neural Networks}
\author{Theogene Habineza$^1$, Ant{\^o}nio H. Ribeiro$^{1}$, Daniel Gedon$^1$,  Joachim A. Behar$^2$, \\ 
 Antonio Luiz P. Ribeiro$^{3}$, Thomas B. Sch{\"o}n$^1$\\
\parbox{0.8\textwidth}{
\begin{flushleft}{\footnotesize \it
$^1$Department of Information Technology, Uppsala University, Sweden\\
$^2$Faculty of Biomedical Engineering, Technion---Israel Institute of Technology, Israel.\\ 
$^3$Department of Internal Medicine, Faculdade de Medicina,  Universidade Federal de Minas Gerais---UFMG,  Brazil}
\end{flushleft}
}}

\date{}

\begin{document}

\maketitle
\begin{abstract}
\textbf{Background:}
Atrial fibrillation (AF) is one of the most common cardiac arrhythmias that affects millions of people each year worldwide and it is closely linked to increased risk of cardiovascular diseases such as stroke and heart failure. Machine learning methods have shown promising results in evaluating the risk of developing atrial fibrillation from the electrocardiogram. We aim to develop and evaluate one such algorithm on a large CODE dataset collected in Brazil.

\textbf{Methods:}
We used the CODE cohort to develop and test a model for AF risk prediction for individual patients from the raw ECG recordings without the use of additional digital biomarkers. The cohort is a collection of ECG recordings and annotations by the Telehealth Network of Minas Gerais, in Brazil. A convolutional neural network based on a residual network architecture was implemented to produce class probabilities for the classification of AF. The probabilities were used to develop a Cox proportional hazards model and a Kaplan-Meier model to carry out survival analysis. Hence, our model is able to perform risk prediction for the development of AF in patients without the condition.

\textbf{Results:}
The deep neural network model identified patients without indication of AF in the presented ECG but who will develop AF in the future with an AUC score of $0.845$. From our survival model, we obtain that patients in the high-risk group (i.e. with the probability of a future AF case being greater than 0.7) are 50\% more likely to develop AF within 40 weeks, while patients belonging to the minimal-risk group (i.e. with the probability of a future AF case being less than or equal to 0.1) have more than 85\% chance of remaining AF free up until after seven years. 

\textbf{Conclusion:}
We developed and validated a model for AF risk prediction. If applied in clinical practice, the model possesses the potential of providing valuable and useful information in decision-making and patient management processes. 

\begin{description}
 \item[Keywords:] Atrial fibrillation; Deep neural network; ECG; Risk prediction; Survival analysis
\end{description}
\end{abstract}

\newpage

\section*{Introduction}

Atrial fibrillation (AF) is progressively more common worldwide within an ageing population \cite{schnabel201550}. It is associated with adverse outcomes such as cognitive impairment and can lead to more severe heart diseases if not treated early. Previous studies have found a close link between AF and increased risk of death~\cite{paixao_evaluation_2020} and heart-related complications, such as stroke and heart failure \cite{litin2018mayo, Odutayoi4482, Wolf1991}. \rvw{R12.}{Good assessment of patient risk can allow more frequent monitoring and facilitate early diagnosis. Early detection of the problem might allow to start} \rvw{R1,1.}{anticoagulation} treatment and help prevent death and disability. 

The electrocardiogram (ECG) is a convenient, fast, and affordable option used at many hospitals, clinics, primary and specialised health centres to diagnose many types of cardiovascular diseases. 
\rvw{R1, 4.}{Over the past 50 years, computer-assisted tools have complemented physician interpretation of ECGs. Notably, the realm of deep learning has emerged as a promising avenue to enhance automated ECG analysis, showcasing impressive strides in recent years \cite{gustafsson3857655artificial, huang2017regularized, ma2017dipole}. Prior studies have predominantly explored the use of deep neural networks (DNNs) to automatically detect AF and other cardiac arrhythmias from standard 12-lead ECGs \cite{attia2019artificial, jeong2021convolutional, ribeiro2020automatic}. This advancement holds valuable implications for clinical decision support, offering auxiliary tools for diagnosing cardiac arrhythmias. However, while achieving consistent diagnoses in patients--even among those with established conditions--is an essential aspect, the parallel need remains for systems yielding timely and early warning for patients with prospective conditions to develop AF.}

Combining the features obtained from DNNs with survival methods is a promising approach for accurate risk prediction. Recent studies explored this approach for the risk prediction of heart diseases \citep{sammani2022life} and mortality \cite{lima2021deep,raghunath2020prediction}. The risk prediction of AF from the 12-lead ECG has been studied before with different approaches and varying degrees of success. \citet{raghunath2021deep} used DNNs for a dataset collected during 30 years to directly predict new-onset AF within one year and identified the patients at risk of AF-related stroke among those predicted to be at high risk of impending AF. The authors in \cite{christopoulos2020artificial} focused on predicting future AF incidents and the time to the event but used a DNN model trained on a different dataset, and the survival analysis spanned a longer period. From our group, \citet{zvuloni2022merging} performed end-to-end AF risk prediction from the 12-lead ECG but did not go further to implement survival modelling and estimate the time to the AF event. \rvw{R1, 3.}{Further}, \citet{biton2021atrial} presents a model that used digital biomarkers in combination with deep representation learning to predict the risk of AF. \rvw{R1, 5.}{Their model uses a random forest classifier including features from a pre-trained DNN where the weights are kept fixed from a different ECG classification task.}

\rvw{R1, 6.}{The aim of our work is to bridge the gap between these studies. While these previous studies focused either on directly predicting future AF cases within a given time frame or incorporated DNNs trained on disparate datasets for survival modelling, there exists no comprehensive approach that synergizes the capabilities of DNNs in AF diagnosis with the precision of survival analysis techniques for estimating time-to-event outcomes. Contrarily, our approach combines both of these aspects: firstly, by employing an end-to-end trained DNN to assess the risk of AF development, and secondly, by utilizing the DNN's output to construct a time-to-event model that forecasts the occurrence of AF from the date of ECG examination. We demonstrate the effectiveness of the method which offers accurate prognostic insights into AF occurrences. Further, we release implementation codes and trained weights to facilitate future studies.}

\section*{Methods}

\subsection*{The dataset}
The model development and testing were conducted using the CODE (Clinical Outcomes in Digital Electrocardiology) dataset \cite{ribeiro2019tele}. The CODE dataset consists of $\num[group-separator={,}]{2322465}$ 12-lead ECG records from $\num[group-separator={,}]{1558748}$ different patients. The ECG records were collected in 811 counties in the state of Minas Gerais, Brazil by a public telehealth system, Telehealth Network of Minas Gerais (TNMG) between 2010 and 2017. A detailed description of the recordings and the labelling process for each ECG exam of the CODE dataset can be found in \cite{ribeiro2020automatic}.

Information about the patients was recorded together with their ECG tracings. The average age of the patients (considering each exam separately) is $53.6$ years with a median of $54$ years and a standard deviation of $17.4$ years. To analyse the natural history of patients with regard to AF, we identified patients who recorded multiple ECG exams. The distribution of the number of visits for each patient during a period of eight years is depicted in Supplementary Material Figure~\ref{fig: visits_count}. As the figure shows, the majority of patients recorded only a single ECG exam ($\num[group-separator={,}]{1104588}$ patients). $\num[group-separator={,}]{285685}$ patients performed two visits each, while the remaining $\num[group-separator={,}]{168475}$ patients recorded ECG exams more than twice. The number of medical visits undertaken by each patient was taken into consideration in classifying the exams into different classes as discussed in the problem formulation.

The ECG signals are between 7 and 10 seconds long and recorded at sampling frequencies ranging from 300 to $600\,\, \text{Hz}$. The ECG records were re-sampled at $400\,\, \text{Hz}$ to generate between 2800 and 4000 temporal samples. All ECGs are zero-padded to obtain a uniform size of 4096 samples for each ECG lead, which are then used as input to the convolutional model.

\rvw{R2 2}{The labels for AF in the CODE dataset were extracted from the text report produced by the expert who looked at the ECGs. To improve the quality of the annotations, some exams were reviewed by doctors, in this case, disagreement with the labels produced by the University of Glasgow automatic diagnosis software was used to select exams to be reviewed. The procedure is described in detail in~\cite{ribeiro2020automatic}.}

\subsection*{Problem formulation}

\begin{figure}[t]
    \centering
    \hspace{200pt}
    \includegraphics[scale=0.425]{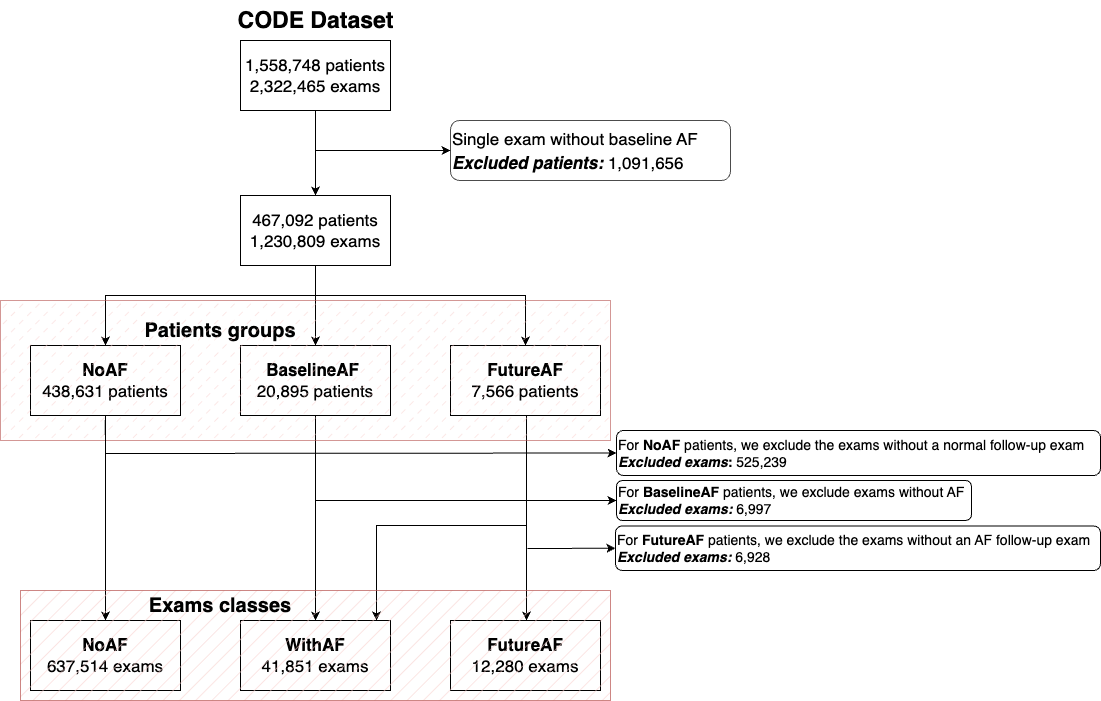}
    \caption{Diagram of patients groups and exams categories.}
    \label{fig: prob_formulation}
\end{figure}

The study considered patients in the CODE database with at least two ECG exams or that have AF. Patients were classified into three groups (NoAF, BaselineAF, FutureAF) according to the presence or absence of a record with AF condition and whether the record with AF is the baseline or not. The ECG exams from the patients were classified into three different classes, focusing on patients who undertook multiple exams. The classification process, which is illustrated in Figure~\ref{fig: prob_formulation}, is detailed as follows:

\begin{itemize}
    \item \textit{NoAF Class:} all ECG exams from patients who recorded multiple exams without presenting an AF abnormality. We exclude the last exam for each patient or exams recorded within one week from the last exam.
    \item \textit{WithAF Class:} combined all ECG exams that exhibit the AF condition.
    \item \textit{FutureAF Class:} regrouped normal ECG exams from patients who had normal ECG exams at the beginning, but who were diagnosed with AF condition in a follow-up exam. The retained records were made before the patients were first diagnosed with AF condition. We exclude all subsequent normal exams after the first positive case, and exams made within one week before this case.
\end{itemize}

\noindent \rvw{R1 7.}{The one-week threshold was set so we don't have to deal with paroxysmal atrial fibrillation cases, which is a brief event of atrial fibrillation that usually stops in 24 hours and may last up to a week. We are interested in using predictions of the FutureAF class for predicting the long-term risk of AF, hence we consider that exams should be distanced by at least one week to be considered as a follow-up exam. Hence, ECG exams recorded within one week before the first exam with the  AF condition were not added FutureAF. Similarly, exams for which we do no follow the patient for longer than one week were not added to NoAF.}

We used the remaining exams for developing and testing the model. In the final dataset, $\num[group-separator={,}]{637514}$ exams ($92.17$\%) belong to the class \textit{NoAF};   $\num[group-separator={,}]{41851}$ ($6.05$\%) to class \textit{WithAF}; and, $\num[group-separator={,}]{12280}$ ($1.78$\%) to the class \textit{FutureAF}. This final dataset was split uniformly at random and by patient into train set, validation set and test set. 60\% of the data were allocated for training, 10\% for validation and 30\% for testing. \rvw{R1, 3.}{Splitting the data into train and validation sets as we have done is common for large datasets such as ours because cross-validation becomes computationally expensive \cite{attia2019artificial,ribeiro2020automatic,hannun2019cardiologist}.} The train-test split happened so that ECG records belonging to one patient ended up in the same split.

\subsection*{DNN architecture and training }
The DNN architecture in this study was based on a deep residual neural network implemented in previous studies~\cite{ribeiro2020automatic, lima2021deep}. The neural network consists of a convolutional layer followed by five residual blocks and ends with a fully connected (dense) layer that passes its output to a softmax to obtain three class probabilities for the classes NoAF, WithAF and FutureAF \rvw{R1, 11.}{which are defined to add up to one}. While the focus is on predicting the class FutureAF from ECG exams with an absence of the AF condition, we kept the exams belonging to the class WithAF to improve the performance of the model. Hence, the developed model also has the capability of conducting automatic AF diagnosis.

The DNN model was trained by minimising the average cross-entropy loss using the Adam optimiser \cite{kingma2015method}. Default parameters were used with weight decay of $5\cdot10^{-4}$ to regularise the model. As the results obtained in \cite{ribeiro2020automatic, lima2021deep} were satisfactory, this study kept most of the selected hyperparameters from these studies. Hence, no further hyperparameter tuning was performed. The initial learning rate was $10^{-3}$ and was reduced by a factor of 10 whenever the validation loss remained without improvement for $7$ consecutive epochs. The dropout rate was manually tuned between values: $0.8$ and $0.5$ with the latter value resulting in improved performance. The training was performed until the minimum learning rate of $10^{-7}$ was reached or for a maximum of 70 epochs. We save and use as the final the one with the best validation results (i.e. minimum error loss) during the optimisation process as a form of early stopping.

\rvw{R1, 3.}{Despite the pronounced class imbalance, we abstain from employing strategies like over- or under-sampling to mitigate it. Over-sampling risks overfitting the minority class, while under-sampling discards numerous majority samples. Since our emphasis lies not on threshold-dependent metrics like accuracy, but rather on utilising the resulting class probabilities for the survival model, the class imbalance becomes less influential.}

\subsubsection*{Model evaluation and metrics}
After the training process, the performance of the DNN model was evaluated on the test data using classification evaluation metrics: sensitivity, positive predictive value (PPV), specificity, false positive rate, $F$-score, the Receiver Operating Characteristic (ROC) curve, Area Under the Receiver Operating Characteristic Curve (AUC-ROC), Precision-Recall Curve and Average Precision (AP) score. This study first evaluated the performance of the model on the task of classifying the three groups: NoAF, WithAF and FutureAF, based on the class probabilities from the DNN model. We plotted the ROC curves, the precision-recall curves and the confusion matrix, and computed the AUC score and AP scores for each class. Next, an evaluation of the model considering only the FutureAF class and the NoAF class was performed to assess the ability of the model to distinguish normal exams within the two classes. In other words, to evaluate how the model performs at AF risk prediction for patients without AF. For this task, samples labelled as WithAF class were removed. The class probabilities for the NoAF class and for the FutureAF class were normalised for each instance to sum to one. Lastly, a probability threshold that maximises the $F_1$-score for NoAF class and FutureAF class was selected, and the threshold-based metrics, namely sensitivity, PPV, specificity and $F_1$-score were computed. The threshold was obtained using the validation set, while all metrics including the plots were measured using the test set.

\subsection*{Time-to-event models}
This study considers non-parametric and semi-parametric methods for time-to-event prediction. Patients in the test set belonging to the class NoAF ($\num[group-separator={,}]{191665}$ recordings, $\num[group-separator={,}]{116255}$ unique patients) and the class FutureAF ($\num[group-separator={,}]{3691}$ recordings, $\num[group-separator={,}]{2016}$ unique patients) were considered for the time-to-event prediction. We used Kaplan-Meier method~\cite{kaplan1958nonparametric} and Cox proportional hazard (PH) models~\cite{cox1972regression}.

\rvw{}{The Kaplan-Meier method~\cite{kaplan1958nonparametric} (also referred to as the product-limit method) is a non-parametric method that provides an empirical estimate of the survival probability at a specific survival time using the actual sequence of the event times. Similar to other non-parametric methods, the advantage of the Kaplan-Meier is that it allows for the analysis without assumptions. On the other hand, the Cox PH model~\cite{cox1972regression} allows us to adjust to different covariates and hence are also interesting to the analysis. Cox PH models are the most commonly used semi-parametric model for survival analysis. The model assumes that the covariates have an exponential influence on the hazard. The log-hazard of an individual is a sum of the population-level baseline hazard and a linear function of the corresponding covariates.}

\rvw{3.2}{We provide two analyses for the Cox PH model, in one analysis we adjust the model with age and gender, and in a second analysis we adjust the model with comorbidities in addition to age and gender. We consider 16 variables that were recorded during a patient visit, that include comorbidities, cardiovascular risk factors and cardiovascular drug usage, namely: use of diuretics, beta-blockers, converting enzyme inhibitors, amiodarone, or calcium blockers, obesity,
diabetes mellitus, smoking, previous myocardial revascularization, family history of coronary heart disease, previous myocardial infarction, dyslipidemia, chronic kidney disease, chronic lung disease, chagas disease, arterial hypertension.} The \textit{observation time} $T$ is given in weeks. During the development of the Cox PH model, patients were subdivided into four groups according to quintiles of the probability output of the DNN: $[0, 0.1)$;  $[0.1, 0.4)$, $[0.4, 0.7)$ and $[0.7, 1.0]$. The study used the first group of patients having a predicted probability of less than \rvw{3.3}{$0.1$} as a reference and produced hazard ratios for the remaining groups. For the Kaplan-Meier model, patients were grouped according to the \rvw{3.3}{same} intervals: $[0, 0.1)$;  $[0.1, 0.4)$, $[0.4, 0.7)$ and $[0.7, 1.0]$. We used the \texttt{lifelines} python library~\cite{Survival-regression}.

\section*{Results}

We developed a model to predict whether a patient belongs to the classes NoAF, WithAF or FutureAF. Our results for the classification task are available in the supplementary material. Since our ultimate goal is to predict the risk of a future AF event, we present here the ability of the model to predict the class FutureAF and the results from survival analysis.

\subsection*{AF risk prediction and survival analysis}
\begin{figure}[!ht]
    \centering
    \includegraphics[scale=0.77]{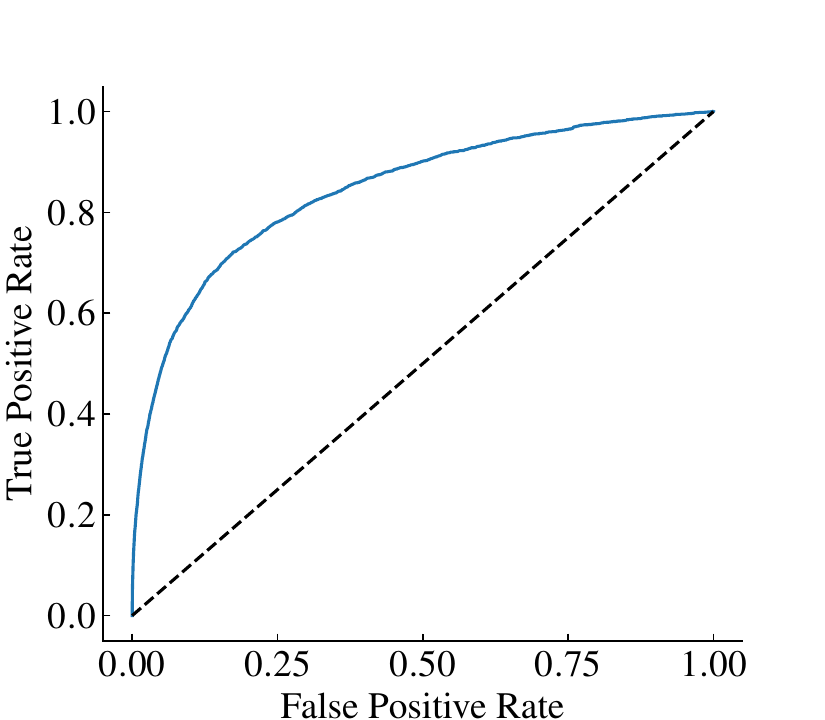}
    \caption{The ROC curves and AUC scores for FutureAF class versus NoAF class. AUC-ROC$=0.845$}
    \label{fig: roc_curve_3vs1}
\end{figure}

\begin{figure}[!ht]
    \centering
    \includegraphics[scale=0.72]{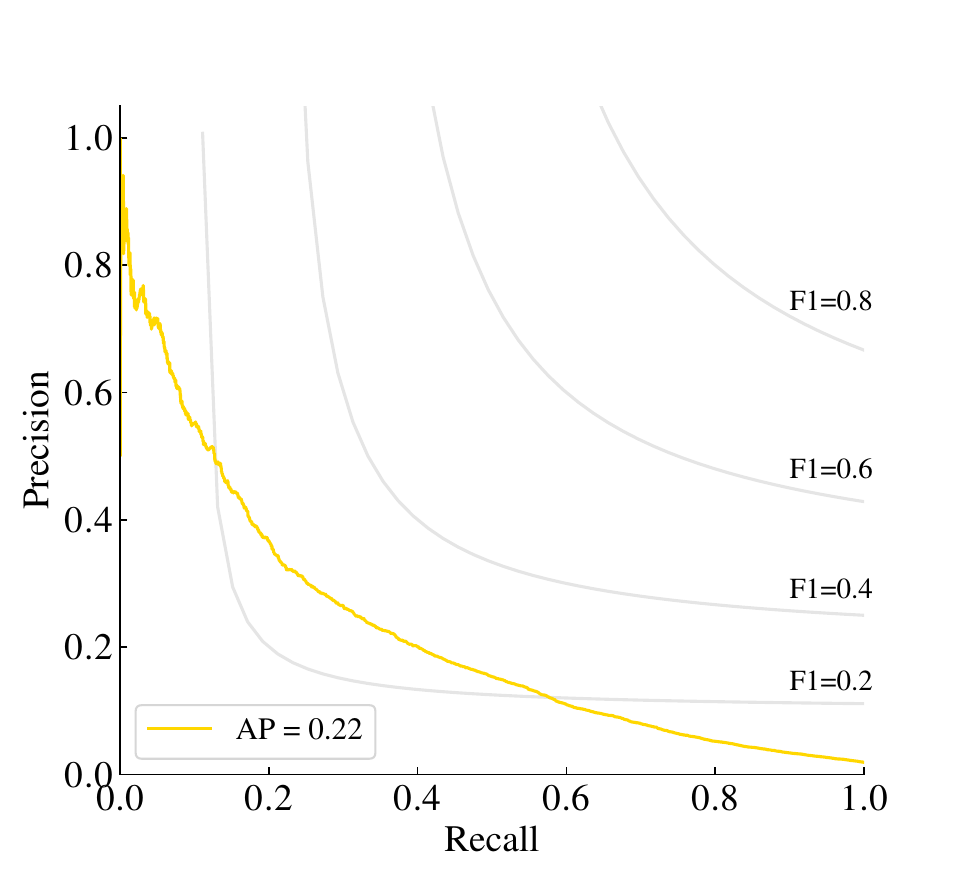}
    \caption{The precision-recall curves and AP scores for FutureAF class versus NoAF class. Recall denotes the sensitivity, and precision denotes the positive predictive value.}
    \label{fig: precision_recall_3vs1}
\end{figure}

The DNN model outputs class probabilities for the three classes. In a first analysis, we excluded exams from the class WithAF in order to study the ability of the model to distinguish between FutureAF and NoAF. We compute the performance metrics using the probability of  FutureAF against that of NoAF. In Table~\ref{tab: confusion_matrix_2classes} we display the confusion matrix, where the predicted values are compared against the true values. In Figure~\ref{fig: roc_curve_3vs1} we show the ROC curve and the AUC-ROC score obtained for this case. The AUC-ROC score was equal to $0.845$. This reveals that the model can detect elements in each class. Figure~\ref{fig: precision_recall_3vs1} displays the PR curves and the calculated average precision (AP) scores. The AP score for the class FutureAF was quite small ($\text{AP} = 0.22$) and its PR curve had a low area under the curve. This suggests that the model is unable to provide both, high sensitivity and PPV values at once for exams in the class FutureAF. 

\begin{table}[!ht]
    \centering
    \caption{Confusion matrix.}
    \small
    \begin{tabularx}{0.4\textwidth}{l l |c c}
    
   & & \multicolumn{2}{c}{\textbf{Predicted Value}} \\
    & &   NoAF   &   FutureAF\\
    \midrule[0,5pt]
    \textbf{True}  &NoAF                &   188\,606  &   3\,059\\
    \textbf{Value}  & FutureAF          &   2\,584    &   1\,107\\
    \bottomrule[0,5pt]
    \end{tabularx}
    \label{tab: confusion_matrix_2classes}
\end{table}

An option for applying the model on the prediction task between two classes is to select a threshold that maximises the $F_1$-score, i.e. putting equal weights on both sensitivity and PPV. The threshold was computed using the validation set and was applied to the classification task for both the validation set and the test set. The obtained optimal probability threshold was equal to $ 0.1043$ and the corresponding performance metrics are shown in Table~\ref{tab: performance_metrics}. All the metrics consider the class FutureAF as the positive class. The sensitivity and PPV values on the test set are $0.322$ and $0.247$, respectively. In contrast, the specificity is very high ($0.981$), which is mainly due to class imbalance. 

\begin{table}[t]
    \centering
    \caption{Performance metrics on the task of predicting the class FutureAF versus NoAF.}\small.
    \small
    \begin{tabularx}{0.457\textwidth}{l c c}
    %Table head
    \toprule[0,5pt]
         & Validation & Test\,\,\,\,(CI $95\%$)\\
    %Table body
    \midrule[0,5pt]
    Sensitivity     & 0.315 		& 0.322\,\,($\pm$ 0.016)\\
    PPV             & 0.250 		& 0.247\,\,($\pm$ 0.012)\\
    Specificity     & 0.982 		& 0.981\,\,($\pm$ 0.001)\\
    F1-score        & 0.279 		& 0.280\,\,($\pm$ 0.012)\\
    \bottomrule[0,5pt]
    \end{tabularx}
    \label{tab: performance_metrics}
\end{table}	

The class probabilities from the DNN model belonging to the class FutureAF were used to develop survival models. Two Cox PH models were implemented, one adjusted with age and gender, and another adjusted with comorbidities in addition to age and gender. Table~\ref{tab: hazard_ratios} shows the hazard ratios of patients whose probabilities for the class FutureAF belong to one of the groups: (0.1-0.4], (0.4-0.7] and (0.7-1.0], taking patients in the group (0.0-0.1] as a reference. As the table indicates, moving from a lower probability range to a higher probability range, the hazards leading to AF also increase. Considering the Cox PH model adjusted with age and gender plus comorbidities, the probability range of (0.7-1.0] had the highest hazard ratio that equals 40.869 (95\% CI: $32.83-50.87; P<0.005$). During the model assessment, however, some covariates (the three probability ranges in this case) did not pass the non-proportional test, hence rejecting the null hypothesis of proportional hazards. This led the study to use a non-parametric model in order to make further survival analyses. A Kaplan-Meier approach was used to this end. 

\begin{table}[t]
    \centering
    \caption{Hazard ratios for different probability groups from the Cox PH model.}
    \small
    \begin{tabularx}{0.805\textwidth}{l cccc}
    \toprule[0,5pt]
    Adjusted for: & Probability Group & Hazard Ratio & CI 95\% & P - value\\
    \midrule[0,5pt]
        & (0.1, 0.4]     & 4.060 & 3.77 - 4.37  &  $<0.005$ \\
    Age and sex & (0.4, 0.7]        & 20.609 & 17.11 - 24.82  & $<0.005$  \\
        & (0.7, 1.0]   & 42.339 & 33.99 - 52.74		  &  $<0.005$ \\
    \midrule[0,5pt]
    Age, sex, risk factors & (0.1, 0.4]     & 3.995 & 3.71 - 4.30  &  $<0.005$ \\
    comorbidities,  & (0.4, 0.7]        & 20.444 & 16.98 - 24.62  & $<0.005$  \\
   \& drug usage$^*$ & (0.7, 1.0]      & 40.869 & 32.83 - 50.87  &  $<0.005$ \\
    \bottomrule[0,5pt]
    \end{tabularx}
    \label{tab: hazard_ratios}\\\vspace{5pt}
    \rvw{}{\footnotesize $^*$\textbf{We adjust for the following comorbidities, cardiovascular risk factors, and drug usage:} use of diuretics, beta-blockers, converting enzyme inhibitors, amiodarone, or calcium blockers, obesity, diabetes mellitus, smoking, previous myocardial revascularization, family history of coronary heart disease, previous myocardial infarction, dyslipidemia, chronic kidney disease, chronic lung disease, chagas disease, arterial hypertension.}
\end{table}	

\begin{figure}[!ht]
    \centering
    \includegraphics[scale=0.852]{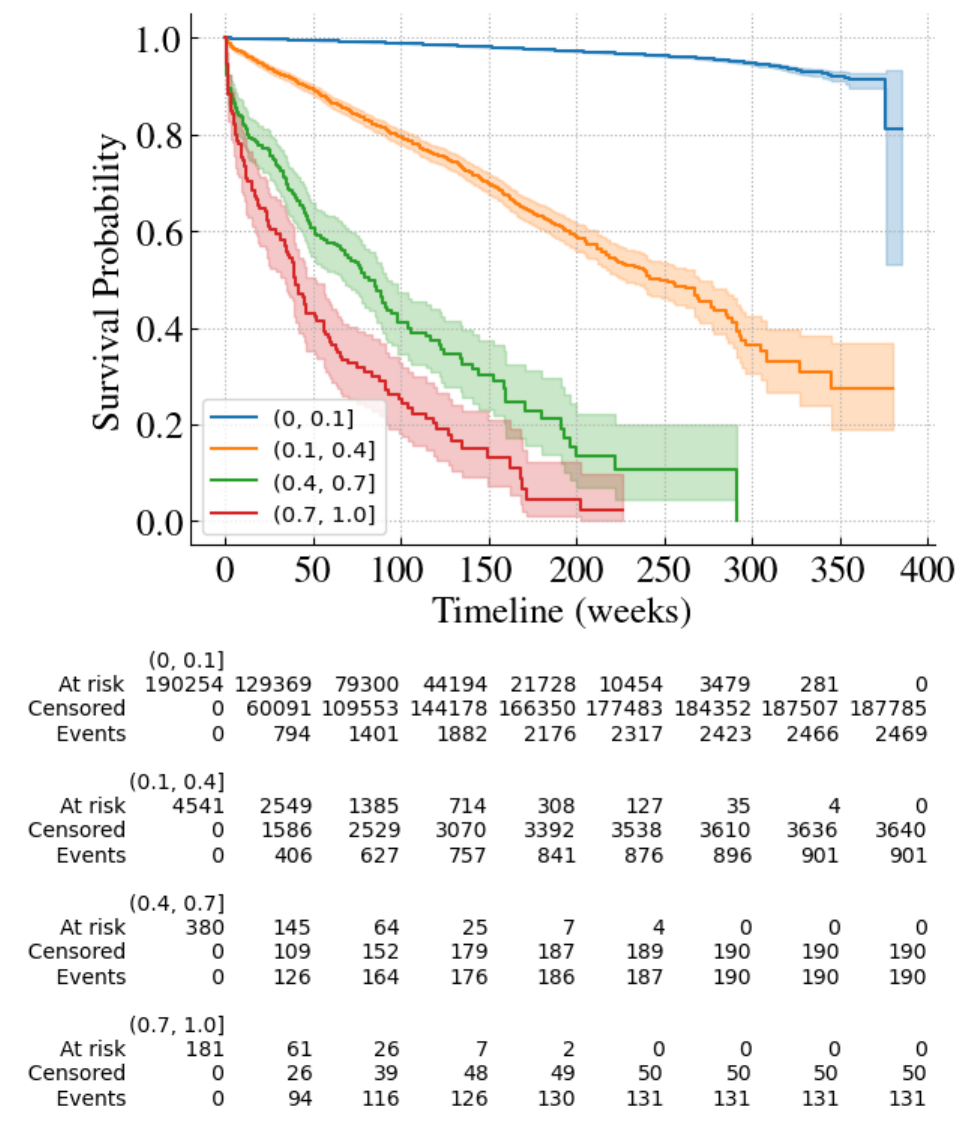}
    \caption{Survival curves for the different cohorts based on their probability range using the Kaplan-Meier model.}
    \label{fig: KM_survival}
\end{figure}

The survival curves that were generated through the Kaplan-Meier estimator are displayed in Figure~\ref{fig: KM_survival}. \rvw{R3,4.}{Note that survival time refers in the context of our study to the time-to-event which is the development of AF and not to actual mortality-related survival.} \rvw{R1,15}{Therefore, survival probability refers to the likelihood that no event occurs.} The shaded area highlights the 95\% confidence interval of the survival probability at different survival times (exponential Greenwood confidence intervals were used \cite{sawyer2003greenwood}). Patients within the lowest risk group maintained survival probabilities greater than $0.8$ during the study period of about seven years. The survival probability is reduced at a higher rate moving from patients in a lower probability range to patients in a higher probability range. The median survival times  for patients in probability groups $(0.0-0.1], \,(0.1-0.4], \,(0.4-0.7]$ and $(0.7-1.0]$ are infinity, 248, 82 and 40 weeks respectively. The median time without developing AF defines the point in time where on average 50\% of the patients in a group would have had the condition. That means for example, patients in the first cohort (probability range $(0.0-0.1]$) have a 50\% chance of never developing AF within seven years, while patients in the last cohort (probability range $(0.7-1.0]$) are 50\% likely to develop AF within 40 weeks (less than a year).

A table below the survival curve in Figure~\ref{fig: KM_survival} shows the number of patients at risk, censored patients (i.e. no further follow-up or the event time is beyond the study period) and patients with AF at different time intervals (50 weeks each time interval). Taking the event times 0 and 50 weeks as an example, for patients within the probability range $(0 - 0.1]$, the number of patients at risk was $\num[group-separator={,}]{129369}\,\, (68\%)$, censored cases were $\num[group-separator={,}]{60091}$ and 794 ($0.42\%$) AF events were recorded after 50 weeks; while for patients within the probability range (0.7 - 1.0] the number of patients at risk was 61 ($33.7\%$), censored cases were 26 and 94 ($51.9\%$) AF events were recorded. This again provides an estimate of the time to event for patients in different risk groups.

\section*{Discussion}

\subsubsection*{DNN model performance}
The DNN model produced a good AUC score for the class FutureAF, which suggests its potential at predicting this class. The actual ability to predict the class FutureAF was attested by the AP score obtained for this class ($\text{AP} = 0.22$). The low score reveals the difficulty in predicting this class and suggests that there would be many false positive cases (incorrectly predicting the class FutureAF) regardless of the threshold.

Regarding the risk prediction task (normal ECG exams in FutureAF vs NoAF), the DNN model produced lower sensitivity and PPV values as shown in Table \ref{tab: performance_metrics} (the probability threshold here maximises $F_1$-score). However, the specificity was as high as $0.982$. This indicates that most of the exams that could be predicted as negative are truly negative and that there would be very few false positive cases. Hence, the information from this prediction task can be of value during a screening of a large population, i.e. one can consider that among the individuals predicted as negative, approximately $1.8\%$ are at risk of developing AF.

\subsubsection*{Survival analysis}
The survival analysis implemented in this study provided additional and valuable information about the risk level and an estimate of the time to the event of having an AF condition. The Cox PH model produced the hazard ratios for patients belonging to four different probability groups taking the group with the lowest risk as a reference. The Cox PH model failed the non-proportional test; still, it provides insight into the risk level incurred by patients in different groups. As stated in \cite{Survival-regression}, a model that does not meet the proportional hazards assumption still can be useful in performing prediction (e.g. predicting survival times) as opposed to making inferences. Recent work also suggests \rvw{}{that} virtually all real-world clinical datasets will violate the proportional hazards assumptions if sufficiently powered and that statistical tests for the proportional hazards assumption may be unnecessary~\cite{stensrud_why_2020}. 

To understand the influence of a class probability group on the survival duration, a Kaplan-Meier model was implemented. The results showed that patients in the highest risk group (FutureAF class probability range of $(0.7 - 1.0]$) were approximately 60\% likely to develop AF within one year, compared with less than 15\% of patients in the minimal risk group (FutureAF class probability range of $(0.0 - 0.1]$) that would develop the condition within the complete time span of seven years. These findings proved the ability of the DNN model at predicting patients with impending AF conditions and with different risk levels. Compared to the results of the study in \cite{biton2021atrial}, which used digital biomarkers from the raw 12-lead ECG, clinical information and features from deep representation learning to make AF risk prediction, our approach learns predicting features directly from the raw ECG signal without the need to extract any biomarker. Thus precluding the need to extract biomarkers from the ECG signal which facilitates the ECG processing pipeline. It is also worth mentioning that the median survival time obtained in \cite{biton2021atrial} is more than two years for patients in probability group $(0.8 - 1.0]$. Even though the methods used to produce survival curves are different (Cox PH model versus Kaplan-Meier) and also the classifier used (Random Forrest versus Neural Network with Softmax), their results seem less alarming considering the results in this work, where $50\%$ of patients in the probability group $(0.7 - 1.0]$ are likely to develop AF within 40 weeks (less than one year). This difference in median survival times may also be attributed to the fact that the study in~\cite{biton2021atrial} used a random forest classifier while this study uses neural networks and a sigmoid function for classification.

\subsubsection*{Clinical implications}
Patients with clinical AF that are not taking anticoagulant medication have an elevated risk of stroke, and the strokes caused by AF are more severe than strokes caused by other causes~\cite{tu_stroke_2015}. AF does not always cause symptoms, and for roughly 20\% of the population, stroke is the first manifestation of AF~\cite{reiffel_incidence_2017}. Thus, there is a lot of interest in detecting cases of AF before the occurrence of a stroke, by systematic screening for asymptomatic AF~\cite{freedman_screening_2017} or, more recently, by the recognition of those in sinus rhythm who will develop AF in the future~\cite{attia2019artificial,zvuloni2022merging, biton2021atrial,alonso_simple_2013,poorthuis_utility_2021}. Among the risk scores that use clinical variables, the CHARGE-AF risk score is one \rvw{}{of} the most accurate and well-validated and uses variables readily available in primary care settings~\cite{alonso_simple_2013}. A recent review of risk scores based on clinical variables for prediction of AF~\cite{poorthuis_utility_2021} found that 14 different scores are potentially useful, with AUC-ROC curves between 0,65 to 0,77 for the general population, with best results for the CHARGE-AF and MHS scores. Risk scores based on standard 12-lead ECGs are a promising tool considering both practical and technical questions~\cite{attia2019artificial,zvuloni2022merging, biton2021atrial}. Reported studies, including ours, showed much higher discrimination capacity, with AUC-ROC curves over 0.85. Since ECGs are routinely performed in most subjects at risk, ie, those older than 60 years old, the prediction can be obtained automatically, without the need of inputting variables in a risk calculator. \rvw{R1 16.}{In this study, we also provide semi-parametric and non-parametric time-to-event models that might help inform doctors of the development of the disease for each group of patients. The model was tested in cases where the disease could be observed up to seven years of the examination, providing a more complete picture for the use of this model in clinical practice}

The ability to accurately recognise patients that have a high chance of developing AF may allow the intensified surveillance of those patients, with early recognition of the appearance of the AF. In this case, the early institution of anticoagulant treatment could prevent the drastic event of a stroke and change the natural history of this condition. Moreover, new therapies to prevent AF could be developed and used for preventing not the stroke but potentially the whole set of complications related to the appearance of AF. All these clinical applications of the method deserve to be tested in controlled clinical trials, but preliminary prospective studies confirmed that AI-augmented ECG analysis could be helpful, at least, to recognise those at higher risk of developing AF~\cite{noseworthy_artificial_2022}.  

\subsubsection*{Limitations}
\rvw{R1 2.1}{One limitation lies in the dataset used for model development and testing. Many of the patients that were considered as all-time normal (without AF during the whole data collection period) had dropped from the follow-up before the study period ended or had a relatively shorter time interval between their first and last ECG records. Therefore, it is impossible to tell with certainty whether an individual was at no risk of developing AF within seven years.} Censored data are unexceptional in survival analysis, however, in normal supervised learning, an ideal dataset would consist of patients who had recorded ECG exams regularly for the considered study period. Moreover, we do not prove this is better than existing clinical scores such as CHARGE-AF~\cite{alonso_simple_2013}.

Similar to a statement in \cite{biton2021atrial}, during data selection, there was a bias towards individuals who had a cardiac disease or a forthcoming heart condition, since all the patients considered had attended multiple medical visits. \rvw{R1 18}{The AF label is also solely based on the ECG analysis. This label might contain errors from medical mistakes and from problems in the extraction of the label (see [11] for a more complete discussion of the labeling process). This way, some FutureAF exams might be previously missed AF cases during the ECG analysis. Finally, the model is developed and tested solely on patients from Brazil, and external validation in other cohorts is needed to verify the efficiency of the model in other populations.}

\section*{Conclusion}

\rvw{R2, 3.}{This study employed ResNet-based convolutional DNNs for end-to-end AF risk prediction from 12-lead ECG signals. The trained DNN effectively identified ECG signal changes indicative of AF development, facilitating risk prediction and survival analysis. By integrating DNN probabilities into Cox PH and Kaplan-Meier models, hazard ratios and survival functions were derived, stratifying patients based on risk levels. This model holds promise for clinical application, aiding AF risk stratification and informing clinical decisions. Further validation is imperative to confirm predictive performance.}

\rvw{R2, 3.}{Future research should encompass external validation on diverse datasets, preferably from distinct geographic populations, to assess model usability across different groups. Exploring the model's potential in identifying AF-related stroke risks is another avenue, considering the established AF-stroke connection \cite{Odutayoi4482, Wolf1991}. Additionally, extending this approach to predict other arrhythmias and cardiovascular diseases is a plausible direction for further development.}

{\small

\subsubsection*{Ethical approval}
This study complies with all relevant ethical regulations. CODE Study was approved by the Research Ethics Committee of the Universidade Federal de Minas Gerais, protocol 49368496317.7.0000.5149. Since this is a secondary analysis of anonymized data stored in the TNMG, informed consent was not required by the Research Ethics Committee for the present study. 

\subsubsection*{Declaration of interests}
There are no competing interests.

\subsubsection*{Funding}
This research is financially supported by the \textit{Wallenberg AI, Autonomous Systems and Software Program (WASP)} funded by Knut and Alice Wallenberg Foundation, and by \emph{Kjell och M{\"a}rta Beijer Foundation}. ALPR is supported in part by CNPq (465518/2014-1, 310790/2021-2 and 409604/2022-4) and by FAPEMIG (PPM-00428-17, RED-00081-16 and PPE-00030-21). ALPR received a Google Latin America Research Award scholarship. JB acknowledges the support of the Technion EVPR Fund: Hittman Family Fund and Grant No ERANET - 3-16881 from the Israeli Ministry of Health. The funders had no role in the study design; collection, analysis, and interpretation of data; writing of the report; or decision to submit the paper for publication.

\subsubsection*{Data sharing}
The DNN model parameters that yield the results from in this paper are available under (\href{https://zenodo.org/record/7038219#.Y9PhldLMJNw}{\url{https://zenodo.org/record/7038219\#.Y9PhldLMJNw}}). 
This should allow the reader to partially reproduce the results from this study. 15\% of the CODE cohort (denoted CODE-15\%) was also made openly available (\href{https://doi.org/10.5281/zenodo.4916206}{\url{https://doi.org/10.5281/zenodo.4916206}}). Researchers affiliated with educational or research institutions might make requests to access the full CODE cohort. Requests should be made to the corresponding author of this paper. They will be forwarded and considered on an individual basis by the Telehealth Network of Minas Gerais. An estimate for the time needed for data access requests to be evaluated is three months. If approved, any data use will be restricted to non-commercial research purposes. The data will only be made available on the execution of appropriate data use agreements.

\subsubsection*{Code availability}
The code for the model training, evaluation and statistical analysis is available at the GitHub repository \href{https://github.com/mygithth27/af-risk-prediction-by-ecg-dnn}{\url{https://github.com/mygithth27/af-risk-prediction-by-ecg-dnn}}.

}

\printbibliography

@inproceedings{litin2018mayo,
  title={Mayo Clinic Family Health Book},
  author={Litin, Scott C and Nanda, Sanjeev},
  year={2018},
  organization={Mayo Clinic}
}

@article{Wolf1991,
    author={Wolf, Philip A and Abbott, Robert D   and Kannel, William B},
    title={Atrial fibrillation as an independent risk factor for stroke: the Framingham Study.},
    journal={Stroke},
    volume={22},
    number={8},
    pages={983-988},
    year={1991},
    doi={10.1161/01.STR.22.8.983},
    URL={https://www.ahajournals.org/doi/abs/10.1161/01.STR.22.8.983},
    eprint={https://www.ahajournals.org/doi/pdf/10.1161/01.STR.22.8.983}
}

@article{Odutayoi4482,
	author={Odutayo, Ayodele and Wong, Christopher X and Hsiao, Allan J and Hopewell, Sally and Altman, Douglas G and Emdin, Connor A},
	title={Atrial fibrillation and risks of cardiovascular disease, renal disease, and death: systematic review and meta-analysis},
	volume={354},
	elocation-id={i4482},
	year={2016},
	doi={10.1136/bmj.i4482},
	publisher={BMJ Publishing Group Ltd},
	URL={https://www.bmj.com/content/354/bmj.i4482},
	eprint={https://www.bmj.com/content/354/bmj.i4482.full.pdf},
	journal={BMJ}
}

@article{ribeiro2020automatic,
  title={Automatic diagnosis of the 12-lead ECG using a deep neural network},
  author={Ribeiro, Ant{\^o}nio H and Ribeiro, Manoel H and Paix{\~a}o, Gabriela MM and Oliveira, Derick M and Gomes, Paulo R and Canazart, J{\'e}ssica A and Ferreira, Milton PS and Andersson, Carl R and Macfarlane, Peter W and Meira Jr, Wagner and Sch{\"o}n, Thomas B and Ribeiro, Antonio Luiz P},
  journal={Nature communications},
  volume={11},
  number={1},
  pages={1--9},
  year={2020},
  publisher={Nature Publishing Group}
}

@article{biton2021atrial,
  title={Atrial fibrillation risk prediction from the 12-lead electrocardiogram using digital biomarkers and deep representation learning},
  author={Biton, Shany and Gendelman, Sheina and Ribeiro, Ant{\^o}nio H and Miana, Gabriela and Moreira, Carla and Ribeiro, Antonio Luiz P and Behar, Joachim A},
  journal={European Heart Journal-Digital Health},
  volume={2},
  number={4},
  pages={576--585},
  year={2021},
  publisher={Oxford University Press}
}

@inproceedings{ma2017dipole,
  title={Dipole: Diagnosis prediction in healthcare via attention-based bidirectional recurrent neural networks},
  author={Ma, Fenglong and Chitta, Radha and Zhou, Jing and You, Quanzeng and Sun, Tong and Gao, Jing},
  booktitle={Proceedings of the 23rd ACM SIGKDD international conference on knowledge discovery and data mining},
  pages={1903--1911},
  year={2017}
}

@article{huang2017regularized,
  title={A regularized deep learning approach for clinical risk prediction of acute coronary syndrome using electronic health records},
  author={Huang, Zhengxing and Dong, Wei and Duan, Huilong and Liu, Jiquan},
  journal={IEEE Transactions on Biomedical Engineering},
  volume={65},
  number={5},
  pages={956--968},
  year={2017},
  publisher={IEEE}
}

@article{ribeiro2019tele,
  title={Tele-electrocardiography and bigdata: the CODE (Clinical Outcomes in Digital Electrocardiography) study},
  author={Ribeiro, Antonio Luiz P and Paix{\~a}o, Gabriela MM and Gomes, Paulo R and Ribeiro, Manoel Horta and Ribeiro, Ant{\^o}nio H and Canazart, Jessica A and Oliveira, Derick M and Ferreira, Milton P and Lima, Emilly M and de Moraes, Jermana Lopes and Castro, Nathalia and Ribeiro, Leonardo B and Macfarlane, Peter W},
  journal={Journal of electrocardiology},
  volume={57},
  pages={S75--S78},
  year={2019},
  publisher={Elsevier}
}

@article{hannun2019cardiologist,
  title={Cardiologist-level arrhythmia detection and classification in ambulatory electrocardiograms using a deep neural network},
  author={Hannun, Awni Y and Rajpurkar, Pranav and Haghpanahi, Masoumeh and Tison, Geoffrey H and Bourn, Codie and Turakhia, Mintu P and Ng, Andrew Y},
  journal={Nature medicine},
  volume={25},
  number={1},
  pages={65--69},
  year={2019},
  publisher={Nature Publishing Group}
}

@article{jeong2021convolutional,
  title={Convolutional Neural Network for Classification of Eight types of Arrhythmia using 2D Time-Frequency Feature Map from Standard 12-Lead Electrocardiogram},
  author={Jeong, Da Un and Lim, Ki Moo},
  journal={Scientific Reports},
  volume={ 11},
  number={20396},
  year={2021},
  URL={https://doi.org/10.1038/s41598-021-99975-6}
}

@article{gustafsson3857655artificial,
  title={Artificial Intelligence-Based ECG Diagnosis of Myocardial Infarction in High-Risk Emergency Department Patients},
  author={Gustafsson, Stefan and Gedon, Daniel and Lampa, Erik and Ribeiro, Ant{\^o}nio H and Holzmann, Martin J and Sch{\"o}n, Thomas B and Sundstrom, Johan},
  journal={Available at SSRN 3857655},
  year={2021},
  URL={http://dx.doi.org/10.2139/ssrn.3857655}
}

@article{cox1972regression,
  title={Regression models and life-tables},
  author={Cox, David R},
  journal={Journal of the Royal Statistical Society: Series B (Methodological)},
  volume={34},
  number={2},
  pages={187--202},
  year={1972},
  publisher={Wiley Online Library}
}

@article{kaplan1958nonparametric,
  title={Nonparametric estimation from incomplete observations},
  author={Kaplan, Edward L and Meier, Paul},
  journal={Journal of the American statistical association},
  volume={53},
  number={282},
  pages={457--481},
  year={1958},
  publisher={Taylor \& Francis}
}

@article{raghunath2020prediction,
  title={Prediction of mortality from 12-lead electrocardiogram voltage data using a deep neural network},
  author={Raghunath, Sushravya and Ulloa Cerna, Alvaro E and Jing, Linyuan and VanMaanen, David P and Stough, Joshua and Hartzel, Dustin N and Leader, Joseph B and Kirchner, Lester H  and Stumpe, Martin C and Hafez, Ashraf and others},
  journal={Nature medicine},
  volume={26},
  number={6},
  pages={886--891},
  year={2020},
  publisher={Nature Publishing Group}
}

@misc{kingma2015method,
  title={A method for stochastic optimization. In: 3rd International Conference on Learning Representations, ICLR 2015, San Diego, CA, USA, May 7-9, 2015, Conference Track Proceedings},
  author={Kingma, Diederik P and Ba, Jimmy},
  year={2015}
}

@misc{Survival-regression,
  author = { Lifelines },
  title = { Survival regression},
  howpublished = {\url{ https://lifelines.readthedocs.io/en/latest/Survival\%20Regression.html}},
  note = {Online; accessed 2.05.2022}
}

@article{lima2021deep,
  title={Deep neural network-estimated electrocardiographic age as a mortality predictor},
  author={Lima, Emilly M and Ribeiro, Ant{\^o}nio H and Paix{\~a}o, Gabriela MM and Ribeiro, Manoel Horta and Pinto-Filho, Marcelo M and Gomes, Paulo R and Oliveira, Derick M and Sabino, Ester C and Duncan, Bruce B and Giatti, Luana and Barreto, Sandhi M. and Meira Jr, Wagner and Thomas B. Sch{\"o}n and Ribeiro, Antonio Luiz P. },
  journal={Nature communications},
  volume={12},
  number={1},
  pages={1--10},
  year={2021},
  publisher={Nature Publishing Group}
}

@article{raghunath2021deep,
  title={Deep neural networks can predict new-onset atrial fibrillation from the 12-lead ECG and help identify those at risk of atrial fibrillation--related stroke},
  author={Raghunath, Sushravya and Pfeifer, John M and Ulloa-Cerna, Alvaro E and Nemani, Arun and Carbonati, Tanner and Jing, Linyuan and vanMaanen, David P and Hartzel, Dustin N and Ruhl, Jeffery A and Lagerman, Braxton F and others},
  journal={Circulation},
  volume={143},
  number={13},
  pages={1287--1298},
  year={2021},
  publisher={Am Heart Assoc}
}

@article{sawyer2003greenwood,
  title={The greenwood and exponential greenwood confidence intervals in survival analysis},
  author={Sawyer, S},
  journal={Applied survival analysis: regression modeling of time to event data},
  pages={1--14},
  year={2003}
}

@article{attia2019artificial,
  title={An artificial intelligence-enabled ECG algorithm for the identification of patients with atrial fibrillation during sinus rhythm: a retrospective analysis of outcome prediction},
  author={Attia, Zachi I and Noseworthy, Peter A and Lopez-Jimenez, Francisco and Asirvatham, Samuel J and Deshmukh, Abhishek J and Gersh, Bernard J and Carter, Rickey E and Yao, Xiaoxi and Rabinstein, Alejandro A and Erickson, Brad J and others},
  journal={The Lancet},
  volume={394},
  number={10201},
  pages={861--867},
  year={2019},
}

@article{zvuloni2022merging,
  title={On Merging Feature Engineering and Deep Learning for Diagnosis, Risk-Prediction and Age Estimation Based on the 12-Lead ECG},
  author={Zvuloni, Eran and Read, Jesse and Ribeiro, Ant{\^o}nio H and Ribeiro, Antonio Luiz P and Behar, Joachim A},
  journal={IEEE Transactions on Biomedical Engineering},
  year={2023},
}

@article{christopoulos2020artificial,
  title={Artificial intelligence--electrocardiography to predict incident atrial fibrillation: a population-based study},
  author={Christopoulos, Georgios and Graff-Radford, Jonathan and Lopez, Camden L and Yao, Xiaoxi and Attia, Zachi I and Rabinstein, Alejandro A and Petersen, Ronald C and Knopman, David S and Mielke, Michelle M and Kremers, Walter and others},
  journal={Circulation: Arrhythmia and Electrophysiology},
  volume={13},
  number={12},
  year={2020},
  publisher={Am Heart Assoc}
}

@article{sammani2022life,
  title={Life-threatening ventricular arrhythmia prediction in patients with dilated cardiomyopathy using explainable electrocardiogram-based deep neural networks},
  author={Sammani, Arjan and van de Leur, Rutger R and Henkens, Michiel THM and Meine, Mathias and Loh, Peter and Hassink, Rutger J and Oberski, Daniel L and Heymans, Stephane RB and Doevendans, Pieter A and Asselbergs, Folkert W and others},
  journal={EP Europace},
  year={2022}
}

@article{schnabel201550,
  title={50 year trends in atrial fibrillation prevalence, incidence, risk factors, and mortality in the Framingham Heart Study: a cohort study},
  author={Schnabel, Renate B and Yin, Xiaoyan and Gona, Philimon and Larson, Martin G and Beiser, Alexa S and McManus, David D and Newton-Cheh, Christopher and Lubitz, Steven A and Magnani, Jared W and Ellinor, Patrick T and others},
  journal={The Lancet},
  volume={386},
  number={9989},
  pages={154--162},
  year={2015},
  publisher={Elsevier}
}

@article{stensrud_why_2020,
  title={Why Test for Proportional Hazards?},
  author={Mats J Stensrud and Miguel A Hernán},
  journal={JAMA},
  pages={1401-1402},
  volume={323},
  year={2023},
}

@article{paixao_evaluation_2020,
  title = {Evaluation of {{Mortality}} in {{Atrial Fibrillation}}: {{Clinical Outcomes}} in {{Digital Electrocardiography}} ({{CODE}}) {{Study}}},
  shorttitle = {Evaluation of {{Mortality}} in {{Atrial Fibrillation}}},
  author = {Paix{\~a}o, Gabriela M. M. and Silva, Luis Gustavo S. and Gomes, Paulo R. and Lima, Emilly M. and Ferreira, Milton P. F. and Oliveira, Derick M. and Ribeiro, Manoel H. and Ribeiro, Antonio H. and Nascimento, Jamil S. and Canazart, J{\'e}ssica A. and Ribeiro, Leonardo B. and Benjamin, Emelia J. and Macfarlane, Peter W. and Marcolino, Milena S. and Ribeiro, Antonio L.},
  year = {2020},
  month = jul,
  journal = {Global Heart},
  volume = {15},
  number = {1},
  pages = {48},
  issn = {2211-8179},
  doi = {10.5334/gh.772},
  urldate = {2020-08-07},
  copyright = {All rights reserved}
}

@article{tu_stroke_2015,
author = {Hans T. H. Tu and Bruce C. V. Campbell and Soren Christensen and Patricia M. Desmond and Deidre A. De Silva and Mark W. Parsons and Leonid Churilov and Maarten G. Lansberg and Michael Mlynash and Jean-Marc Olivot and Matus Straka and Roland Bammer and Gregory W. Albers and Geoffrey A. Donnan and Stephen M. Davis},
title ={Worse Stroke Outcome in Atrial Fibrillation is Explained by More Severe Hypoperfusion, Infarct Growth, and Hemorrhagic Transformation},
journal = {International Journal of Stroke},
volume = {10},
number = {4},
pages = {534-540},
year = {2015},
doi = {10.1111/ijs.12007},
    note ={PMID: 23489996},

URL = { 
        https://doi.org/10.1111/ijs.12007
    
},
eprint = { 
        https://doi.org/10.1111/ijs.12007
    
}
}

@article{reiffel_incidence_2017,
    author = {Reiffel, James A. and Verma, Atul and Kowey, Peter R. and Halperin, Jonathan L. and Gersh, Bernard J. and Wachter, Rolf and Pouliot, Erika and Ziegler, Paul D. and for the REVEAL AF Investigators},
    title = "{Incidence of Previously Undiagnosed Atrial Fibrillation Using Insertable Cardiac Monitors in a High-Risk Population: The REVEAL AF Study}",
    journal = {JAMA Cardiology},
    volume = {2},
    number = {10},
    pages = {1120-1127},
    year = {2017},
    month = {10},
    issn = {2380-6583},
    doi = {10.1001/jamacardio.2017.3180},
    url = {https://doi.org/10.1001/jamacardio.2017.3180},
    eprint = {https://jamanetwork.com/journals/jamacardiology/articlepdf/2650790/jamacardiology\_reiffel\_2017\_oi\_170047.pdf},
}

@article{freedman_screening_2017,
author = {Ben Freedman  and John Camm  and Hugh Calkins  and Jeffrey S. Healey  and Mårten Rosenqvist  and Jiguang Wang  and Christine M. Albert  and Craig S. Anderson  and Sotiris Antoniou  and Emelia J. Benjamin  and Giuseppe Boriani  and Johannes Brachmann  and Axel Brandes  and Tze-Fan Chao  and David Conen  and Johan Engdahl  and Laurent Fauchier  and David A. Fitzmaurice  and Leif Friberg  and Bernard J. Gersh  and David J. Gladstone  and Taya V. Glotzer  and Kylie Gwynne  and Graeme J. Hankey  and Joseph Harbison  and Graham S. Hillis  and Mellanie T. Hills  and Hooman Kamel  and Paulus Kirchhof  and Peter R. Kowey  and Derk Krieger  and Vivian W. Y. Lee  and Lars-Åke Levin  and Gregory Y. H. Lip  and Trudie Lobban  and Nicole Lowres  and Georges H. Mairesse  and Carlos Martinez  and Lis Neubeck  and Jessica Orchard  and Jonathan P. Piccini  and Katrina Poppe  and Tatjana S. Potpara  and Helmut Puererfellner  and Michiel Rienstra  and Roopinder K. Sandhu  and Renate B. Schnabel  and Chung-Wah Siu  and Steven Steinhubl  and Jesper H. Svendsen  and Emma Svennberg  and Sakis Themistoclakis  and Robert G. Tieleman  and Mintu P. Turakhia  and Arnljot Tveit  and Steven B. Uittenbogaart  and Isabelle C. Van Gelder  and Atul Verma  and Rolf Wachter  and Bryan P. Yan  and A Al Awwad  and F Al-Kalili  and T Berge  and G Breithardt  and G Bury  and WR Caorsi  and NY Chan  and SA Chen  and I Christophersen  and S Connolly  and H Crijns  and S Davis  and U Dixen  and R Doughty  and X Du  and M Ezekowitz  and M Fay  and V Frykman  and M Geanta  and H Gray  and N Grubb  and A Guerra  and J Halcox  and R Hatala  and H Heidbuchel  and R Jackson  and L Johnson  and S Kaab  and K Keane  and YH Kim  and G Kollios  and ML Løchen  and C Ma  and J Mant  and M Martinek  and I Marzona  and K Matsumoto  and D McManus  and P Moran  and N Naik  and T Ngarmukos  and D Prabhakaran  and D Reidpath  and A Ribeiro  and A Rudd  and I Savalieva  and R Schilling  and M Sinner  and S Stewart  and N Suwanwela  and N Takahashi  and E Topol  and S Ushiyama  and N Verbiest van Gurp  and N Walker  and T Wijeratne },
title = {Screening for Atrial Fibrillation},
journal = {Circulation},
volume = {135},
number = {19},
pages = {1851-1867},
year = {2017},
doi = {10.1161/CIRCULATIONAHA.116.026693},

URL = {https://www.ahajournals.org/doi/abs/10.1161/CIRCULATIONAHA.116.026693},
eprint = {https://www.ahajournals.org/doi/pdf/10.1161/CIRCULATIONAHA.116.026693}
}

@article{alonso_simple_2013,
author = {Alvaro Alonso  and Bouwe P. Krijthe  and Thor Aspelund  and Katherine A. Stepas  and Michael J. Pencina  and Carlee B. Moser  and Moritz F. Sinner  and Nona Sotoodehnia  and João D. Fontes  and A. Cecile J. W. Janssens  and Richard A. Kronmal  and Jared W. Magnani  and Jacqueline C. Witteman  and Alanna M. Chamberlain  and Steven A. Lubitz  and Renate B. Schnabel  and Sunil K. Agarwal  and David D. McManus  and Patrick T. Ellinor  and Martin G. Larson  and Gregory L. Burke  and Lenore J. Launer  and Albert Hofman  and Daniel Levy  and John S. Gottdiener  and Stefan Kääb  and David Couper  and Tamara B. Harris  and Elsayed Z. Soliman  and Bruno H. C. Stricker  and Vilmundur Gudnason  and Susan R. Heckbert  and Emelia J. Benjamin },
title = {Simple Risk Model Predicts Incidence of Atrial Fibrillation in a Racially and Geographically Diverse Population: the CHARGE-AF Consortium},
journal = {Journal of the American Heart Association},
volume = {2},
number = {2},
pages = {e000102},
year = {2013},
doi = {10.1161/JAHA.112.000102},

URL = {https://www.ahajournals.org/doi/abs/10.1161/JAHA.112.000102},
eprint = {https://www.ahajournals.org/doi/pdf/10.1161/JAHA.112.000102}

}

@article{poorthuis_utility_2021,
	title = {Utility of risk prediction models to detect atrial fibrillation in screened participants},
	volume = {28},
	issn = {2047-4881},
	doi = {10.1093/eurjpc/zwaa082},
	language = {eng},
	number = {6},
	journal = {European Journal of Preventive Cardiology},
	author = {Poorthuis, Michiel H. F. and Jones, Nicholas R. and Sherliker, Paul and Clack, Rachel and de Borst, Gert J. and Clarke, Robert and Lewington, Sarah and Halliday, Alison and Bulbulia, Richard},
	month = may,
	year = {2021},
	pmid = {33624100},
	pmcid = {PMC8651014},
	pages = {586--595},
}

@article{noseworthy_artificial_2022,
	title = {Artificial intelligence-guided screening for atrial fibrillation using electrocardiogram during sinus rhythm: a prospective non-randomised interventional trial},
	volume = {400},
	issn = {0140-6736},
	url = {https://www.sciencedirect.com/science/article/pii/S0140673622016373},
	doi = {https://doi.org/10.1016/S0140-6736(22)01637-3},
	number = {10359},
	journal = {The Lancet},
	author = {Noseworthy, Peter A and Attia, Zachi I and Behnken, Emma M and Giblon, Rachel E and Bews, Katherine A and Liu, Sijia and Gosse, Tara A and Linn, Zachery D and Deng, Yihong and Yin, Jun and Gersh, Bernard J and Graff-Radford, Jonathan and Rabinstein, Alejandro A and Siontis, Konstantinos C and Friedman, Paul A and Yao, Xiaoxi},
	year = {2022},
	pages = {1206--1212},
}
\newpage 

\section*{Supplementary material}

\pagenumbering{roman} 

% Reset page
\setcounter{page}{1}

% Reset 
\setcounter{equation}{0}
\renewcommand{\theequation}{S.\arabic{equation}}%

\setcounter{figure}{0}
\renewcommand{\thefigure}{S.\arabic{figure}}%

\setcounter{table}{0}
\renewcommand{\thetable}{S.\arabic{table}}%

\subsection*{Automatic AF diagnosis}
\begin{figure}[!ht]
    \centering
    \includegraphics[scale=0.72]{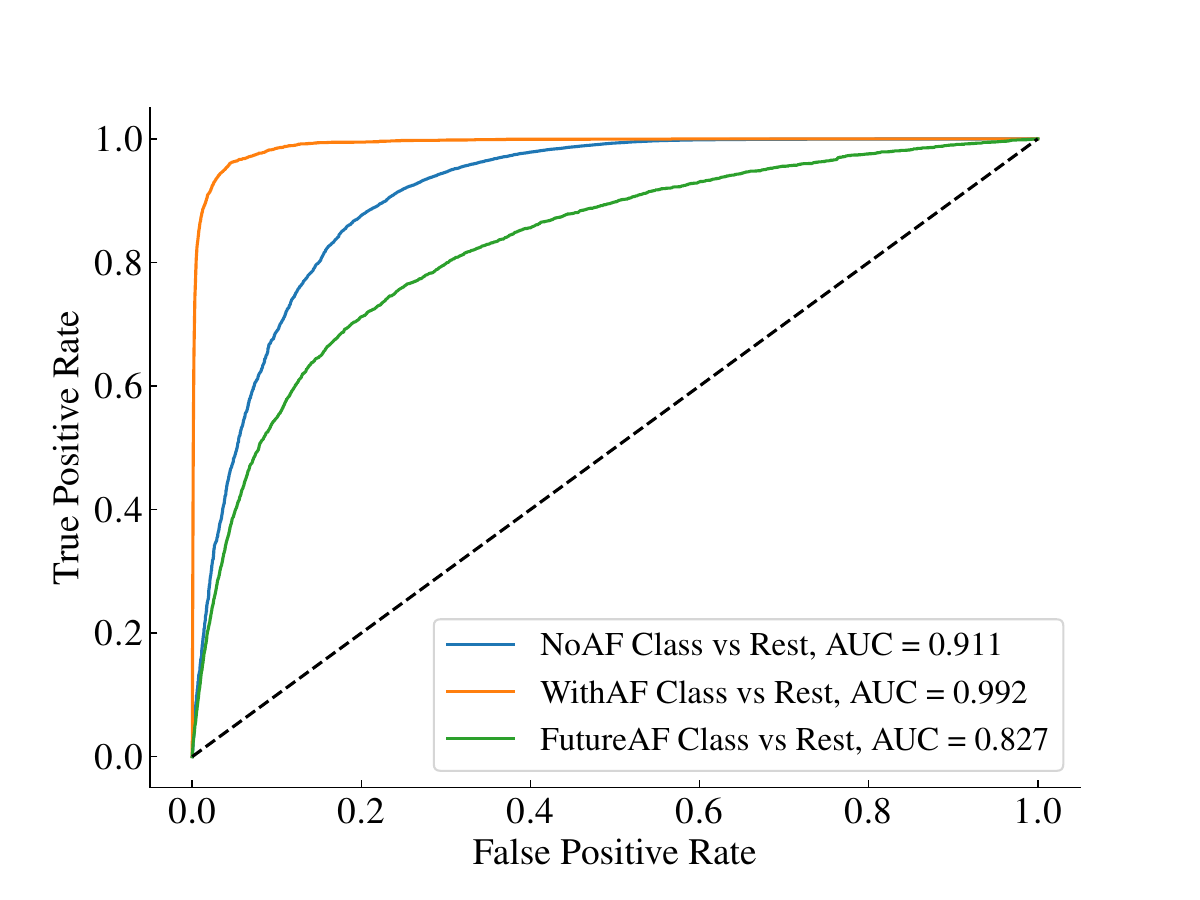}
    \caption{The ROC curves and AUC scores for the three classes.}
    \label{fig: multiclass_roc}
\end{figure}

We developed a model to predict whether a patient belongs to the classes NoAF, WithAF or FutureAF. Supplementary Figure~\ref{fig: multiclass_roc} displays the ROC curves obtained by considering each class against the other two classes. The highest obtained AUC score was at predicting the class WithAF versus the rest $(\text{AUC} = 0.992)$, followed by predicting the class NoAF versus the rest $(\text{AUC} = 0.911)$ and predicting the class FutureAF versus the rest $(\text{AUC} = 0.827)$. The AUC score for the class WithAF is nearly optimal. 

\begin{figure}[!ht]
    \centering
    \includegraphics[scale=0.78]{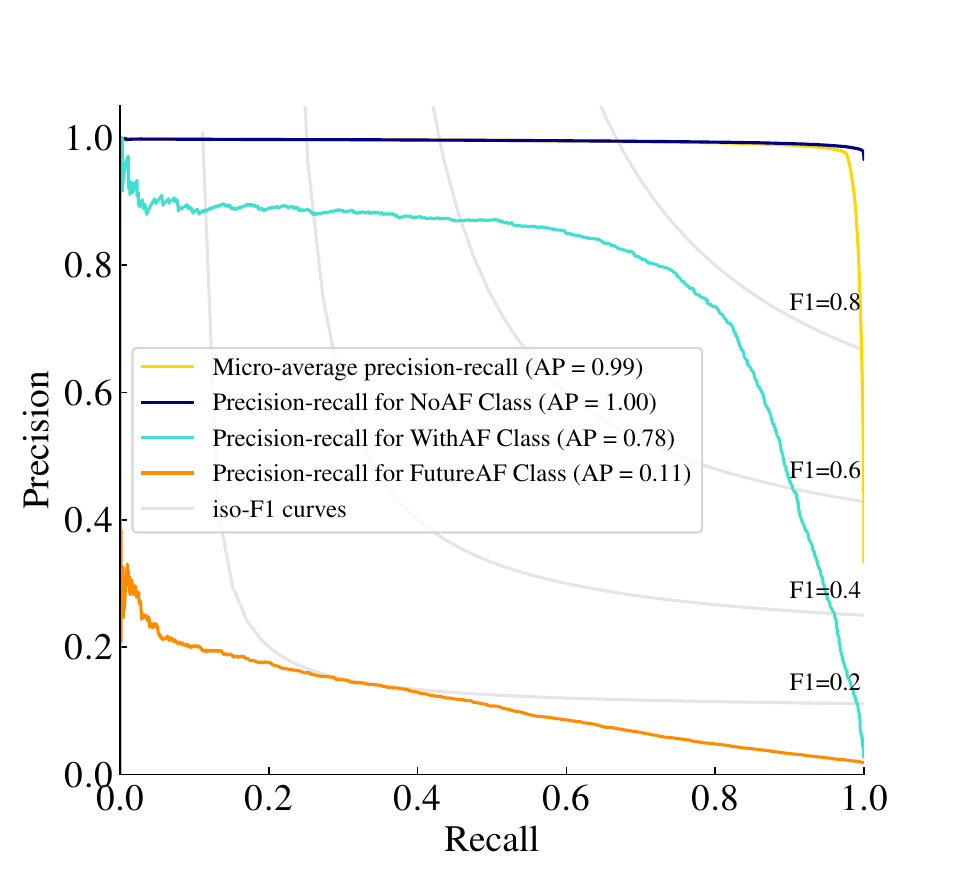}
    \caption{The precision-recall curves and AP scores for the three classes. Recall denotes the sensitivity, and precision denotes the positive predictive value.}
    \label{fig: precision_recall_curve}
\end{figure}

Supplementary Figure~\ref{fig: precision_recall_curve} displays the precision-recall (PR) curves and the average precision (AP) score for each class. In addition, a micro-average PR curve that gives a general view of the performance of the model by considering all the classes was presented. The PR curve for the class NoAF had the highest area under the curve and an AP score close to 1. This was expected as the majority of ECG exams belonged to class NoAF ($92.17\%$ of the whole dataset for model development). The higher AP score for the class NoAF was also reflected in the micro-average AP score, which was high as well. The PR curve for the class WithAF had a lower area under the curve and an AP score equalling $0.78$. This AP score also confirms the ability of the model at distinguishing abnormal ECG exams from normal ones. The model produced a much lower area under the curve and a low AP score ($\text{AP} = 0.11$) for the class FutureAF.  

\begin{figure}[hbt!]
    \centering
    \includegraphics[scale=0.779]{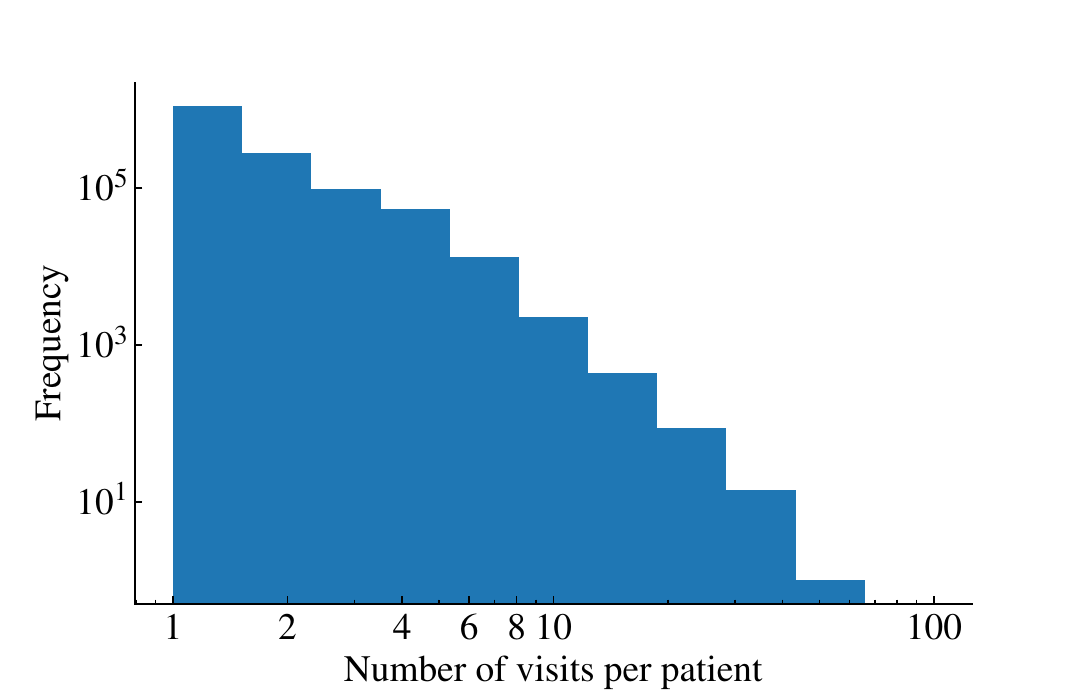}
    \caption{A distribution of the number of medical visits per patient.} 
    \label{fig: visits_count}
\end{figure}

\end{document}